\documentclass[10pt,conference,a4paper]{IEEEtran}
\IEEEoverridecommandlockouts
\usepackage{graphicx}
\usepackage{epstopdf}
\usepackage{amsmath,amsmath,amssymb}
\usepackage{mathrsfs}
\usepackage{bm} 
\usepackage{multirow}
\usepackage{cite}
\usepackage{algorithmic}
\usepackage{textcomp}
\usepackage{xcolor}

\def\BibTeX{{\rm B\kern-.05em{\sc i\kern-.025em b}\kern-.08em
		T\kern-.1667em\lower.7ex\hbox{E}\kern-.125emX}}

\begin{document}

\title{Prior Knowledge about Attributes:\\Learning a More Effective Potential Space for Zero-Shot Recognition}
\author{\IEEEauthorblockN{Chunlai Chai}
	\IEEEauthorblockA{\textit{Zhejiang Gongshang University} \\
		Hangzhou, China \\
		ccl@mail.zjgsu.edu.cn}
	\and
	\IEEEauthorblockN{Yukuan Lou}
	\IEEEauthorblockA{\textit{Zhejiang Gongshang University} \\
		Hangzhou, China \\
		louyukuan@gmail.com}
	\and
	\IEEEauthorblockN{Shijin Zhang}
	\IEEEauthorblockA{\textit{Zhejiang Gongshang University} \\
		Hangzhou, China \\
		zhangshijin0304@gmail.com}
	\and
	\IEEEauthorblockN{Ming Hua}
	\IEEEauthorblockA{\textit{Oakland University} \\
		State of Michigan, US \\
		ming@oakland.edu}
}
\maketitle
\begin{abstract}
Zero-shot learning (ZSL) aims to recognize unseen classes accurately by learning seen classes and known attributes, but correlations in attributes were ignored by previous study which lead to classification results confused. To solve this problem, we build an Attribute Correlation Potential Space Generation (ACPSG) model which uses a graph convolution network and attribute correlation to generate a more discriminating potential space. Combining potential discrimination space and user-defined attribute space, we can better classify unseen classes. Our approach outperforms some existing state-of-the-art methods on several benchmark datasets, whether it is conventional ZSL or generalized ZSL.
\end{abstract}


\section{Introduction}
Recently, recognizing objects by training neural networks is becoming more important because of deep neural networks \cite{1} \cite{2} \cite{3} \cite{4}. However the deeper and larger the neural network, the more data and labels it needs to be trained. In other words, the success of large neural networks is inextricably linked to the large amount of manually collected labeled data \cite{5}. This appears to be inconsistent with "intelligence". One of the goals of our machine learning aims to reduce labor costs. This question has aroused the interest of many researchers, so zero-shot learning (ZSL) \cite{6} \cite{7} \cite{8} and few-shot learning (FSL) came into being.

ZSL is a more drastic scheme, and its goal is to learn from seen classes so as to achieve the classification of unseen classes. Different from the traditional machine learning classification problem, the target class of ZSL is never seen in training. We associate the unseen and seen classes through auxiliary semantic attributes. What’s more, Learning of seen classes and their auxiliary semantic properties, model learned how to use the semantic attribute space predicts unseen classes and infers their labels by searching for the class with the most similar semantic attributes.

In general, the typical scheme of the most advanced ZSL methods is (1) to extract the feature representation for seen data from CNN models pre-trained on the large-scale datasets, (2) to learn mapping functions to project the visual features and semantic attribute representations to potential space, (3) using generative models, e.g. GAN \cite{24}, VAE \cite{25}, generates "fake" features about unseen classes, thus transforming ZSL problems into ordinary supervised learning problems.

\begin{figure}[]
	\centering
	\includegraphics[height=6.8cm,width=9cm]{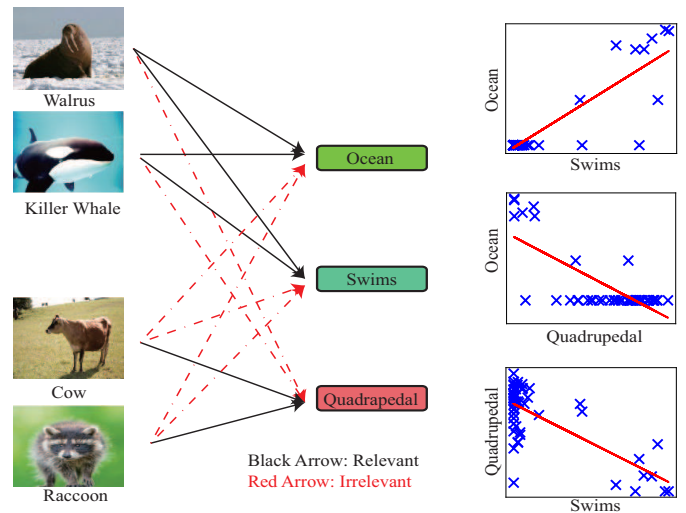}
	\caption{An illustrative diagram of semantic attributes correlation. It can be seen from the figure that if an animal has a certain attribute, it is likely to also have an attribute related to it; otherwise, it is likely not to have an attribute not related to it.}
	\label{figl}
\end{figure}

During the training phase, zero-shot learning can be divided into inductive ZSL \cite{8} \cite{9} \cite{10} \cite{11} \cite{12} \cite{13} and transduced ZSL \cite{14} \cite{15} \cite{16} \cite{17} \cite{18} \cite{19}. Inductive ZSL has no visual information and no semantic attribute information of the seen classes during training, while the transforming ZSL can access some unlabeled images. In the test phase, two types of ZSL which are conventional ZSL \cite{20} and generalized ZSL \cite{21}. In conventional ZSL, the images in the test phase are only from the unseen classes, for generalized ZSL, the images in the test phase can be from the unseen and seen classes.

But in previous studies, the correlation between semantic attributes has not been explored. All semantic attributes are considered as independent, and they do not affect each other. In fact, semantic attributes are related. For example, the correlation among three attributes in Fig. \ref{figl}. We can find that walruses and killer whales have the attributes of swims and ocean, but they do not have the attributes of quadrupeds. Corresponding to this, cows and raccoons with quadruped attributes while do not have swims and ocean attributes. The occurrence of this situation is consistent with the correlation among attributes swims, ocean, and quadrupedal. In terms of attributes, swims are positively related to the ocean, and quadrupeds are negatively related to the ocean and swims. It is clearly that if we can explore the relationship in attributes, we can perform well in the recognition and classification of object. We believe that ignore this will lead to ambiguity in the semantic attribute space.

We solve this problem by graph convolutional networks (GCN) \cite{22}, GCN uses the correlation between class nodes and semantic attribute nodes to generate a latent space to help identify unseen classes. Before this research, the bipartite graph has been used to represent the correlation between ZSL nodes, as it is shown in Fig. \ref{fig2}, but the bipartite graph ignores the correlation in semantic attribute nodes. We propose a new graph model to replace the bipartite graph, covering the correlation between semantic attribute nodes, thus generating better potential space. A new ZSL framework called Attribute Correlation Potential Space Generation (ACPSG) model consists of two parts, the first part generates latent discrimination attribute space from GCN, the second part maps the visual features of the unseen classes into user-defined attribute space and latent discrimination attribute space through an autoencoder \cite{23}. In the end, combining multiple spaces, we can consider both the UA and LA spaces to perform ZSL prediction.

In summary,

(1) We first propose the attribute correlation in ZSL and applied it to the inductive ZSL framework.

(2) We carried out a lot of experiments and analysis on three zero-shot learning datasets and proved the superior performance of our proposed method in this regard.

\section{Related Work}
\subsection{Zero-Shot Learning}
According to previous research \cite{26}, there are three types of models in ZSL: (1) Class-Inductive Instance-Inductive setting, it means training the model using only the trainable instances and the set of seen labels, (2) Class-Transductive Instance-Inductive setting, This means to train models using trainable instances and a set of seen labels, as well as a set of unseen labels. (3) Class-Transductive Instance-Inductive setting, it means to train the model using trainable instances and seen label sets, as well as unseen label sets and corresponding unlabeled test sets.

Several methods work better now. Firstly, Based on Generative Adversarial Networks \cite{27}. This model uses the generator to generate "fake" unseen class features, and then train the classifier to learn these features to complete the classification. In essence, it turns the ZSL problem into a common supervised learning problem. Secondly, Based on Variational Auto-encoder \cite{49}. This model uses two VAEs with the same structure which are to encode the image and decode the class embedding. Finally, Based on Zoom Network \cite{29}. Extract the key areas of the image through a zoom net and enlarge the key areas, at the same time, CNN is used to train the original image and key areas to achieve classification.

Besides, some researchers have proposed general-purpose ZSL model research, trying to use different forms of source data to build the model, such as using text/audio to build the model, which is also a topic of the current academe. 

\subsection{Graph Neural Networks}
Graph convolution \cite{22} was first proposed to extend CNN to graphs and to directly process graph-type data. CNN generally acts on Euclidean space, and cannot directly act on non-Euclidean space. Many important data sets are stored in the form of graphs in reality, such as social network information, knowledge map, protein network, the World Wide Web and so on. The form of these graph networks is not like an image. It is composed of a neatly arranged matrix but is unstructured information. CNN cannot be used for feature extraction, but graph convolution can be applied here.

The core of graph convolution is that each node in the graph is affected by neighbor nodes and further points at any time, so it constantly changes its state until the final balance. The nodes closer to the target node have a greater influence on the target node. GCN has subtly designed a method for extracting features from graph data so that we can use these features to perform node classification, graph classification, and edge prediction on graph data. It is versatile that we can get embedded representations of graphs in this way.

\subsection{AutoEncoder}
An autoencoder \cite{30} is an unsupervised neural network model. It can learn the hidden features of the input data, which can be called encoding. At the same time, the new input features can be used to reconstruct the original input data, which is called decoding. Intuitively, auto-encoders can be used to reduce the feature dimension is similar to principal component analysis, but its performance is stronger than PCA. This is because neural network models can extract more efficient new features. In addition to feature dimensionality reduction, new features learned by the autoencoder can be input into a supervised learning model, so the autoencoder can be used as a feature extractor. As an unsupervised learning model, autoencoders can also be used to generate new data that is different from the training samples, such as variational autoencoders \cite{31}.

\section{OUR MODEL}
\subsection{Notation and Problem Formulaiton}
We will specify the mathematical notation as follows: a seen dataset $\mathcal{S}=\{({\boldsymbol x}_i^s,{\boldsymbol y}_i^s)\}_{i=1}^{N^s}$,where $N^s$ represents the number of seen samples, each one ${\boldsymbol x}_i^s$ represents a seen sample, and ${\boldsymbol y}_i^s\in\mathcal{Y}^\mathcal{S}$ is the corresponding label. a unseen dataset $\mathcal{U}=\{({\boldsymbol x}_i^u,{\boldsymbol y}_i^u)\}_{i=1}^{N^u}$,where $N^u$ represents the number of unseen samples,each one ${\boldsymbol x}_i^u$ represents a unseen sample, and ${\boldsymbol y}_i^u\in\mathcal{Y}^\mathcal{U}$ is the corresponding label. The dataset $\mathcal{S}$ is used as the training set and $\mathcal{U}$ is used as the testing set. The seen and unseen classes are disjoint, i.e., $\mathcal{Y}^\mathcal{S}\cap\mathcal{Y}^\mathcal{U}=\varnothing$,$\mathcal{Y}^\mathcal{S}\cup\mathcal{Y}^\mathcal{U}=\mathcal{Y}$. $\boldsymbol X^s=\theta(\boldsymbol x^s)$ denotes the visual features, and $\boldsymbol Y^s=\phi(\boldsymbol y^s)$ denotes the attributes bound to $\boldsymbol X^s$. $\boldsymbol X^u=\theta(\boldsymbol x^u)$ denotes the visual features, and $\boldsymbol Y^u=\phi(\boldsymbol y^u)$ denotes the attributes bound to $\boldsymbol X^u$.

Overall, the goal of ZSL is to optimize the following functions:
\begin{equation}\label{1}
\mathcal{F}=\frac{1}{N^s}\sum_{i=1}^{N^s}||\phi(\boldsymbol y_i^s)-\phi(\emph{f}(\boldsymbol x_i^s,\boldsymbol W))||^2+\lambda\Omega(\boldsymbol W)
\end{equation}
where $\emph{f}$ is a mapping function that maps samples to corresponding labels, $\boldsymbol W$ representing parameters in the mapping process, and $\Omega$ is the regular term, which is used to prevent overfitting. In this paper, we express $\emph{f}$ as follows:
\begin{equation}\label{2}
\emph{f}(\boldsymbol x_i^s,\boldsymbol W)=\mathop{\arg\min}_{\boldsymbol y\in\mathcal{Y}}\mathcal{D}(\boldsymbol W^T\theta(\boldsymbol x_i^s),\phi(\boldsymbol y))
\end{equation}
where $\mathcal{D}$ is a distance function. $\theta$ refers to the image feature extractor, e.g., AlexNet \cite{32}, GoogleNet \cite{33}, RestNet \cite{34} etc. $\phi(\boldsymbol y)$ represents the relevant attributes of $\boldsymbol y$.

\subsection{Attribute Correlation}

\begin{figure}[]
	\centering
	\includegraphics[height=6cm,width=9cm]{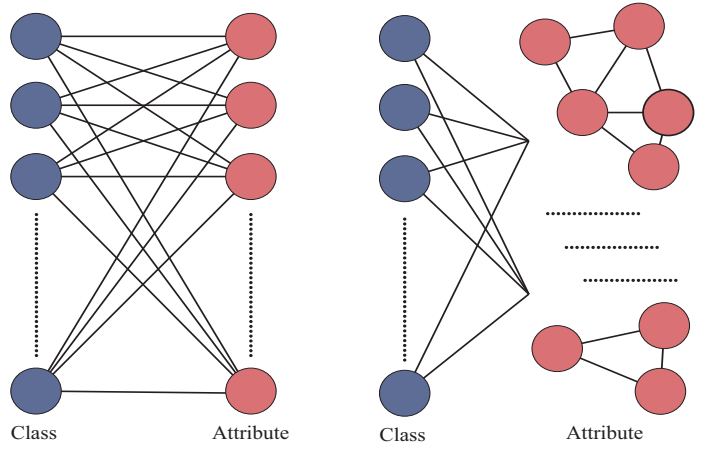}
	\caption{An illustrative diagram of the correlation between nodes. The figure on the left shows the connection between nodes in the past ZSL research, i.e., Bipartite Graph. The figure on the right shows the connection between nodes in our research.}
	\label{fig2}
\end{figure}

In the past research, attributes were considered as independent and unrelated individuals, but we tried to find the correlation between the attributes and used this correlation as a prior knowledge to find effective potential space. We use the following form to represent the correlation between classes and attributes:
\begin{equation}\label{3}
\boldsymbol C=[\phi(\boldsymbol y_1^s),...,\phi(\boldsymbol y_{N^s}^s),\phi(\boldsymbol y_{N^s+1}^u),...,\phi(\boldsymbol y_{N^s+N^u}^u)]^T\in \mathbb{R}^{d_C \times d_T}
\end{equation}
where $d_C$ represents the number of classes and $d_T$ represents the number of attributes. We use graph $\mathcal{G}=(\mathcal{V},\mathcal{E})$ to represent the correlation between various nodes in ZSL, $\mathcal{V}$ and $\mathcal{E}$ separately refer to the node-set and the edge-set, as shown in Fig. \ref{fig2}, every edge in the graph describes the interaction between nodes and nodes, and in mathematically we use the adjacency matrix to represent the correlation of nodes in the graph.

Now we introduce the covariance matrix to describe the connection between attributes:
\begin{equation}\label{4}
\begin{aligned}
\boldsymbol W_{ij}&=\emph{COV}(\phi(\boldsymbol y_i),\phi(\boldsymbol y_j))\\
&=\emph{E}[(\phi(\boldsymbol y_i)-\boldsymbol \mu_i)(\phi(\boldsymbol y_j)-\boldsymbol \mu_j)]\\
\boldsymbol W&=\emph{COV}(\boldsymbol C) \in \mathbb{R}^{d_T \times d_T}
\end{aligned}
\end{equation}
where $\boldsymbol \mu$ is the mean. For calculation convenience, we scale the range of $\boldsymbol W$ to be consistent with $\boldsymbol C$. By calculating the covariance, we obtained the correlation between the attributes from known prior knowledge. Now we can use the adjacency matrix $\boldsymbol A$ to describe the node correlations in the graph $\mathcal{G}$:
\begin{equation}\label{5}
\boldsymbol A=\begin{bmatrix} \boldsymbol 0_{d_C}&\boldsymbol C\\\boldsymbol C^T&\boldsymbol W \odot (\boldsymbol \sim \boldsymbol I)
\end{bmatrix}\boldsymbol \in \mathbb{R}^{(d_C+d_T) \times (d_C+d_T)}
\end{equation}
where $\boldsymbol I$ represents the identity matrix, $\boldsymbol \sim$ represents take inverse, and $\odot$ represents the matrix dot product. Each element $\boldsymbol A_{ij}$ in the adjacency matrix $\boldsymbol A$ represents the correlation between the $\boldsymbol i$ node and the $\boldsymbol j$ node in the graph $\mathcal{G}$.

\subsection{Latent Space Generation}
Inspired by \cite{22} \cite{35}, we use graph convolutional networks to generate potential discernment spaces. 

Due to the abundant attribute information, we might as well regard $\boldsymbol F$ as the node features:
\begin{equation}\label{6}
\boldsymbol F=\begin{bmatrix} \boldsymbol 0_{d_C}&\boldsymbol C\\\boldsymbol C^T&\boldsymbol 0_{d_T}
\end{bmatrix}\boldsymbol \in \mathbb{R}^{(d_C+d_T) \times (d_C+d_T)}
\end{equation}

Besides, let the diagonal matrix $\boldsymbol D$ denote degree matrix with $\boldsymbol D_{ii}=d_i=\sum_j \boldsymbol A_{ij}$ and $\boldsymbol S=\boldsymbol D^{-\frac{1}{2}}\boldsymbol A\boldsymbol D^{-\frac{1}{2}}$ is the normalized adjacency matrix.

To use the correlation between nodes more effectively, we consider the following diffusion function \cite{36}:

\begin{equation}\label{7}
\mathcal{O}(\boldsymbol H)=\sum_{i,j} \frac{\boldsymbol A_{ij}}{2}||\frac{\boldsymbol H_i}{\sqrt{d_i}}-\frac{\boldsymbol H_j}{\sqrt{d_j}}||^2+\mu\sum_i||\boldsymbol H_i-\boldsymbol F_i||^2
\end{equation}
\begin{figure*}[]
	\centering
	\includegraphics[height=11cm,width=13cm]{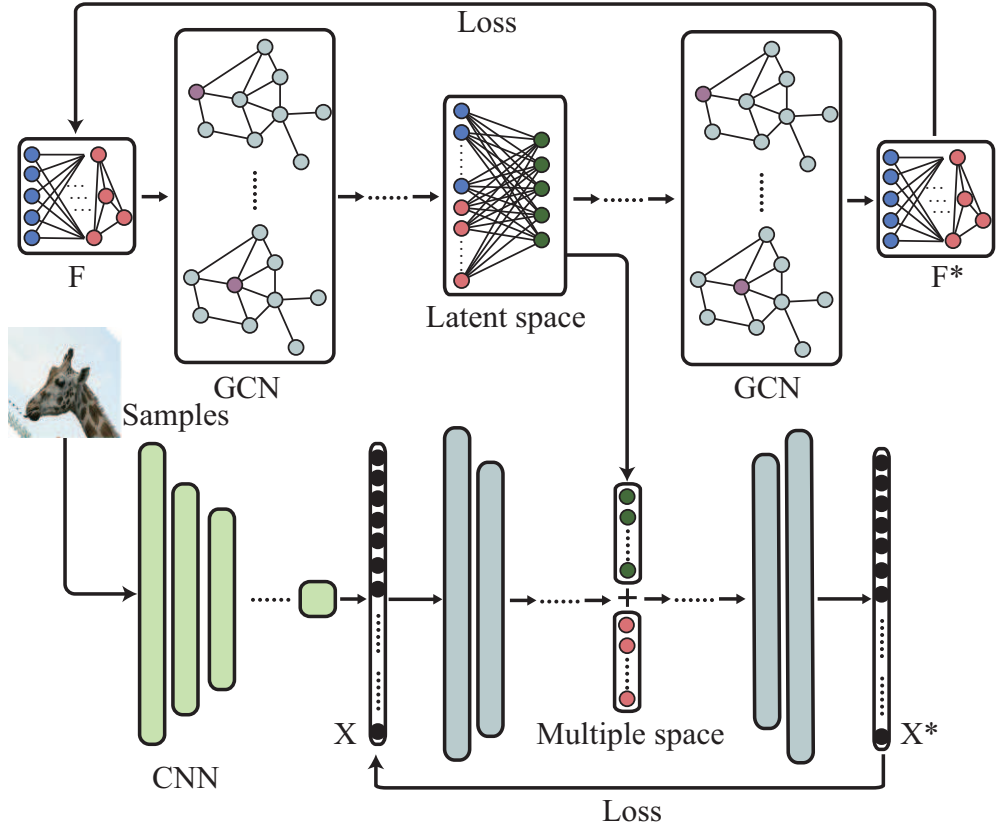}
	\caption{Overall illustration of the framework proposed in this paper. At the first stage, we added the correlation between attributes as a prior knowledge, using a graph convolution model to generate a latent discernment space. In the second stage, we use autoencoders to map visual features into multiple spaces and learn a reliable decoder.}
	\label{fig3}
\end{figure*}
where $\boldsymbol H_i$ represents the $i$-th node mapped in the latent space. Essentially, the first term implies that the information flows along high weight edges, which forces a node similar to its neighbors. In contrast, the second term tends to maintain the original features, that is, to preserve the global information.The coefficient $\mu>0$ constrains the balance between these two terms. Furthermore, the optimal solution is:

\begin{equation}\label{8}
\begin{aligned}
&\boldsymbol H^* \propto (\boldsymbol{I}-\alpha \boldsymbol{S})^{-1}\boldsymbol{F}\\
&\alpha=\frac{1}{1+\mu}
\end{aligned}
\end{equation}

Due to $\alpha \boldsymbol{S} \in [0,1)$, it is natural to generalize the Eq. \ref{8} to a higher-order form:
\begin{equation}\label{9}
\boldsymbol H^* \propto \sum_{k=0}^{\infty} (\alpha\boldsymbol{S})^k \boldsymbol{F}
\end{equation}

To avoid calculating infinite terms and overfitting, we consider the suggestion of \cite{35} to truncate $k$. In spectral graph theory \cite{22}, the convolution operation on the graph is formulated as:
\begin{equation}\label{10}
\boldsymbol{g}_\theta \star \boldsymbol{F}=\boldsymbol{V}\boldsymbol{G}_\theta(\Lambda)\boldsymbol{V}^T\boldsymbol{F}
\end{equation}
where $\boldsymbol{g}_\theta$ and $\boldsymbol{G}_\theta(\Lambda)$ separately denote the spatial filter and the spectral filter, and $\boldsymbol{V}$ is the eigenvectors of the normalized Laplacian matrix $\boldsymbol{L}=\boldsymbol{I}-\boldsymbol{S}=\boldsymbol{V}\Lambda\boldsymbol{V}^T$. In \cite{35}, the reported graph convolution operation $\boldsymbol{g}_\theta \star \boldsymbol{F}$ can be finally expressed as:
\begin{equation}\label{11}
\boldsymbol{g}_\theta \star \boldsymbol{F}=\sum_{k=0}^{p} (\alpha\boldsymbol{S})^k \boldsymbol{F}
\end{equation}

We can see that Eq. \ref{9} and Eq. \ref{11} have similar expressions. As shown in Fig. \ref{fig3}, The truncated graph convolution operation is applied to the autoencoder, which can generate potential discrimination space. This process can be expressed as:
\begin{equation}\label{12}
\boldsymbol{H}^{(l+1)}=\sigma(\sum_{k=0}^{p}(\alpha \boldsymbol{S})^k\boldsymbol{H}^{(l)}\boldsymbol{W}^{(l)})
\end{equation}
where $\boldsymbol{H}^{(l)}$ is the $l$-th layer of the network,
$\boldsymbol{H}^{(0)}=\boldsymbol{F}$, and $\boldsymbol{W}^{(l)}$ is the $l$-th weight of network. The dimension of the latent discrimination space generated after the encoding phase is $\mathbb{R}^{(d_C+d_T)\times d}$. We use $\psi(\boldsymbol{y})$ to represent the latent discrimination attribute of class $\boldsymbol{y}$.

\subsection{Visual Feature Mapping}
With the learned latent discrimination space, the goal of our framework is to map the visual features to user-defined attribute space and latent discrimination attribute space, i.e., multiple spaces. We use the following loss function for optimization:
\begin{equation}\label{13}
\mathcal{L}_{en}=||\boldsymbol{W}_{en}\boldsymbol{X}^s-[\phi(\boldsymbol{y}^s);\psi(\boldsymbol{y}^s)]||
\end{equation}
where $\boldsymbol{W}_{en}$ is the parameter used in encoding, essentially, $\boldsymbol{W}_{en}$ is a parameterized matrix used to measure the similarity between $\boldsymbol{X}^s$ and multiple space. And $[\phi(\boldsymbol{y}^s);\psi(\boldsymbol{y}^s)]\in\mathbb{R}^{(d_T+d)\times N_s}$ is multiple space attribute. Similarly, at the decoding stage, we also map the attributes of the multiple space back to visual features and constrain them:
\begin{equation}\label{14}
\mathcal{L}_{de}=||\boldsymbol{W}_{de}\boldsymbol{M}-\boldsymbol{X}^s||
\end{equation}
where $\boldsymbol{M}=\boldsymbol{W}_{en}\boldsymbol{X}^s$. With described above, we minimize the following loss function:
\begin{equation}\label{15}
\mathcal{L}=\mathcal{L}_{en}+\lambda\mathcal{L}_{de}
\end{equation}
where $\lambda_1$ is weighting coefficients. As shown in Fig. \ref{fig3}, the model can learn a robust encoder $\boldsymbol{W}_{en}$. Finally, the labels to identify unseen classes can be recognition by following equation:
\begin{equation}\label{16}
\boldsymbol{y}^*=\mathop{\arg\min}_{\boldsymbol y\in\mathcal{Y}}\mathcal{D}(\boldsymbol W_{en}\boldsymbol{X},[\phi(\boldsymbol{y});\psi(\boldsymbol{y})])
\end{equation}
where $\mathcal{D}$ is a distance function.

\section{EXPERIMENTS}
To evaluate the effectiveness of ACPSG, we conducted sufficient experiments on several general-purpose datasets. At the same time, ablation experiments were also performed to verify the validity of the theory.
\subsection{Datasets and Setting}
$\bullet\quad$Animals with Attributes 2 (AwA2) \cite{5}: This dataset provides a platform to benchmark transfer-learning algorithms, in particular attribute base classification and zero-shot learning. It consists of 37322 images of 50 animal classes.

$\bullet\quad$Caltech UCSD Birds 200 (CUB) \cite{37}: This dataset is all pictures of birds, with a total of 200 classes, 150 classes for the training set, and 50 classes for the test set. The semantics of the class is 312 dimensions and there are 11,788 pictures.

$\bullet\quad$Attribute Pascal and Yahoo (aPY) \cite{38}: This data set has a total of 32 classes, 20 classes are used as training sets and 12 classes are used as test sets. The class semantics are 64 dimensions and there are 15339 pictures in total.
\begin{table}[h]
	\centering
	\caption{A DETAILED DESCRIPTION OF THE DATASETS.}
\begin{tabular}{||c|c|c|c|c||}
	\hline
	Dataset&Attribute&Sample&Seen/Unseen&Dim\\
	\hline
	AWA2&85&37322&40/10&40\\
	\hline
	CUB&312&11788&150/50&40\\
	\hline
	aPY&64&18627&20/12&20\\
	\hline
\end{tabular}
\end{table}

Table 1 gives the relevant information on the above three data sets. In this paper, a pre-trained model ResNet101 \cite{39} is used to extract the feature values of the picture. The generated feature values are 2048-dimensional vectors. Dim represents the dimension of the latent discrimination space. Related information can be queried by \cite{5}.

We implemented our framework with Pytorch. In the phase of generating potential discernment space, we set $p=2$ and $\alpha=0.8$. In the visual feature mapping phase, we set $\lambda=1$. We use Adam \cite{40} to optimize our model. Distance function we choose cosine function.

\begin{figure}
	\centering
	\includegraphics[height=5cm,width=9cm]{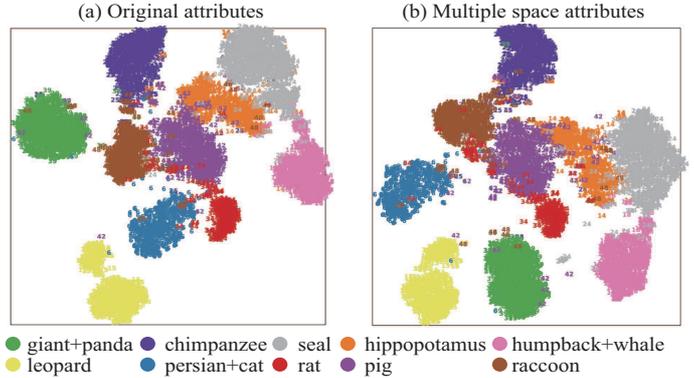}
	\caption{The t-SNE \cite{48} visualization of multiple space on AwA2. (a) represents the predicted attribute distribution of the unseen classes without using potential discrimination space. (b) represents the predicted attribute distribution of unseen classes in the case of multiple spaces. Different colors represent different species.}
	\label{fig4}
\end{figure}

\subsection{Evaluation Metrics}
To evaluate the performance of ACPSG, we use the following two methods:

(1)Use the encoder to map visual features to multiple spaces as semantic attributes, and then calculate the distance between the semantic attributes and class prototypes:
\begin{equation}\label{17}
\boldsymbol{y}^*=\mathop{\arg\min}_{\boldsymbol y\in\mathcal{Y}}\mathcal{D}(\boldsymbol W_{en}\boldsymbol{X},[\phi(\boldsymbol{y});\psi(\boldsymbol{y})])
\end{equation}
where $\boldsymbol{y}^*$ the class label of the sample, and $\mathcal{D}$ is the distance function.

(2)Use a decoder to map each class prototype to visual semantic space as a visual feature, and then calculate the distance between them and the sample visual feature:
\begin{equation}\label{18}
\boldsymbol{y}^*=\mathop{\arg\min}_{\boldsymbol y\in\mathcal{Y}}\mathcal{D}(\boldsymbol{W}_{de}[\phi(\boldsymbol{y});\psi(\boldsymbol{y})],\boldsymbol{X})
\end{equation}
where $\boldsymbol{y}^*$ the class label of the sample, and $\mathcal{D}$ is the distance function.

We use $\mathcal{Y}^t$ to represent the target class to be detected. Under conventional ZSL,  $\mathcal{Y}^t=\mathcal{Y}^u$ refers to searching only in the unseen label set. But in generalized ZSL everything is different, $\mathcal{Y}^t=\mathcal{Y}^u\cup\mathcal{Y}^s=\mathcal{Y}$, it means that the class to be searched has expanded from the unseen class to the entire class. So we use the harmonic mean$(\mathcal{H})$ to measure the performance of the model under generalized ZSL:
\begin{equation}\label{19}
\emph{H}=\frac{2\times\emph{ACC}_s\times\emph{ACC}_u}{\emph{ACC}_s+\emph{ACC}_u}
\end{equation}
where $\emph{ACC}_s$ denotes the accuracy on the seen test class and $\emph{ACC}_u$ denotes the accuracy on the unseen test class.

\subsection{Conventional Zero-Shot Learning}
In conventional ZSL, we first generate a potential discrimination space, and then map the visual features into the potential discrimination space and the user-defined space, at last, we can consider both the user-defined 
\begin{table*}[h]
	\centering
	\caption{THE EXPERIMENTAL RESULTS ON THE CONVENTIONAL ZSL. HERE THE PS AND THE SS SEPARATELY REFER TO THE PROPOSED SPLIT AND THE STANDARD SPLIT. THE BEST RESULT IS MARKED IN BOLD FONT. NONE MEANS NO POTENTIAL DISCERNMENT SPACE IS USED. $S->V$ MEANS USE EQ. \ref{18} TO CALCULATE ACCURACY, $V->S$ MEANS USE EQ. \ref{17} TO CALCULATE ACCURACY.}
	\setlength{\tabcolsep}{8mm}{
		\begin{tabular}{||l|l|l|l||}
			\hline
			\multicolumn{1}{||c|}{\multirow{2}{*}{\textbf{Method}}} &
			\multicolumn{1}{c|}{\textbf{AwA2}} &
			\multicolumn{1}{c|}{\textbf{CUB}} &
			\multicolumn{1}{c||}{\textbf{aPY}} \\ \cline{2-4} 
			\multicolumn{1}{||c|}{} &
			SS\qquad \quad PS &
			SS\qquad \quad PS &
			SS\qquad \quad PS \\ \hline
			\begin{tabular}[c]{@{}l@{}}DAP \cite{41}\\ CONSE \cite{42}\\ ALE \cite{6}\\ ESZSL \cite{10}\\ SJE \cite{7}\\ SYNC \cite{43}\\ SAE \cite{23}\\ SE-ZSL \cite{16}\\ ZSKL \cite{28}\\ F-CLSWGAN \cite{17}\\ DCN \cite{44}\\ PSRZSL \cite{45}\end{tabular} &
			\begin{tabular}[c]{@{}l@{}}58.7\qquad 46.1\\ 67.9\qquad 44.5\\ 80.3\qquad 62.5\\ 75.6\qquad 58.6\\ 69.5\qquad 61.9\\ 71.2\qquad 46.6\\ 80.2\qquad 54.1\\ 80.8\qquad 69.2\\ - \quad \qquad $\boldsymbol{70.2}$\\ -\qquad \qquad -   \\ -\qquad \qquad  -   \\ - \quad \qquad 63.8\end{tabular} &
			\begin{tabular}[c]{@{}l@{}}37.5\qquad 40.0\\ 36.7\qquad 34.3\\ 53.2\qquad 54.9\\ 55.1\qquad 53.9\\ 55.3\qquad 53.9\\ 54.1\qquad 55.6\\ 33.4\qquad 33.3\\ 60.3\qquad 59.6\\ - \quad \qquad 57.1\\ - \quad \qquad 61.5\\ 55.6\qquad 56.2\\ - \quad \qquad 56.0\end{tabular} &
			\begin{tabular}[c]{@{}l@{}}35.2\qquad 33.8\\ 25.9\qquad 26.9\\ - \quad \qquad 39.7\\ 34.4\qquad 38.3\\ 32.0\qquad 32.9\\ 39.7\qquad 23.9\\ $\boldsymbol{55.4}$\qquad 8.3\\ - \qquad \qquad  -\\ - \quad \qquad $\boldsymbol{45.3}$\\ - \qquad \qquad  -\\ - \quad \qquad 43.6\\ - \quad \qquad  38.4\end{tabular} \\ \hline
			\begin{tabular}[c]{@{}l@{}}None (S-\textgreater{}V)\\ ACPSG (S-\textgreater{}V)\\ None (V-\textgreater{}S)\\ ACPSG (V-\textgreater{}S)\end{tabular} &
			\begin{tabular}[c]{@{}l@{}}78.1\qquad 57.9\\ 79.5\qquad 66.1\\ 79.8\qquad 48.9\\ $\boldsymbol{82.8}$\qquad 49.8\end{tabular} &
			\begin{tabular}[c]{@{}l@{}}59.2\qquad 58.6\\ 61.3\qquad 59.1\\ 58.1\qquad 57.4\\ $\boldsymbol{62.7}$\qquad $\boldsymbol{60.4}$\end{tabular} &
			\begin{tabular}[c]{@{}l@{}}41.7\qquad 23.4\\ 45.1\qquad 27.3\\ 41.3\qquad 24.4\\ 47.2\qquad 27.5\end{tabular} \\ \hline
	\end{tabular}}
\end{table*}
\begin{table*}[h]
	\centering
	\caption{THE EXPERIMENTAL RESULTS ON THE GENERALIZED ZSL. $S$ REPRESENTS THE ACCURACY OF SEEN CLASSES. $U$ REPRESENTS THE ACCURACY OF UNSEEN CLASSES. $H$ IS THE HARMONIC MEAN. THE BEST RESULT IS MARKED IN BOLD FONT.}
	\setlength{\tabcolsep}{6mm}{
		\begin{tabular}{||l|l|l|l||}
			\hline
			\multicolumn{1}{||c|}{\multirow{2}{*}{\textbf{Method}}} & \multicolumn{1}{c|}{\textbf{AwA2}} & \multicolumn{1}{c|}{\textbf{CUB}} & \multicolumn{1}{c||}{\textbf{aPY}} \\ \cline{2-4} 
			\multicolumn{1}{||c|}{}                                 & $S$\qquad $U$\qquad $H$                    & $S$\qquad $U$\qquad $H$                   & $S$\qquad $U$\qquad $H$                   \\ \hline
			\begin{tabular}[c]{@{}l@{}}CONSE \cite{42}\\ CMT \cite{46}\\ SJE \cite{7} \\ ESZSL \cite{10}\\ SYNC \cite{43}\\ SAE \cite{23}\\ LATEM \cite{47}\\ ALE \cite{6}\\ ZSKL \cite{28}\\ PSRZSL \cite{45}\\ DCN \cite{44}\end{tabular} &
			\begin{tabular}[c]{@{}l@{}}$\boldsymbol{90.6}$ \, 0.5 \quad 1.0\\ 90.0 \quad  0.5 \quad 1.0\\ 73.9 \quad  8.0 \quad 14.4\\ 77.8 \quad  5.9 \quad 11.0\\ 90.5 \quad 10.0 \, 18.0\\ 82.2 \quad  1.1 \,\quad 2.2\\ 77.3 \quad 11.5 \, 20.0\\ 81.8 \quad 14.0  \, 23.9\\ 82.7 \quad 18.9 \, 30.8\\ 73.8 \quad 20.7 \, 32.3\\ -  \qquad  -  \qquad  -\end{tabular} &
			\begin{tabular}[c]{@{}l@{}}$\boldsymbol{72.2}$ \,  1.6 \quad  3.1\\ 49.8 \quad  7.2 \quad  12.6\\ 59.2 \quad  23.5 \, 33.6\\ 63.8 \quad  12.6  \, 21.0\\ 70.9  \quad 11.5 \,  19.8\\ 54.0 \quad  7.8  \quad 13.6\\ 57.3  \quad 15.2 \, 24.0\\ 62.8 \quad  23.7\quad 34.4\\ 52.8  \quad 21.6 \quad 30.6\\ 54.3 \quad  24.6  \quad 33.9\\ 37.0 \quad  $\boldsymbol{25.5}$  \, 30.2\end{tabular} &
			\begin{tabular}[c]{@{}l@{}}$\boldsymbol{91.2}$  \quad 0.0  \quad 0.0\\ 74.2 \quad  10.9 \,  19.0\\ 55.7 \quad  3.7 \quad  6.9\\ 70.1  \quad 2.4  \quad 4.6\\ 66.3 \quad  7.4 \quad  13.3\\ 80.9 \quad  0.4  \quad 0.9\\ 73.0 \quad  0.1 \quad  0.2\\ 73.7  \quad 4.6  \quad 8.7\\ 76.2 \quad  10.5 \,  18.5\\ 51.4 \quad  13.5 \,  21.4\\ 75.0 \quad  $\boldsymbol{14.2}$ \ $\boldsymbol{23.9}$\end{tabular} \\ \hline
			Ours                                                   & 82.5 \quad $\boldsymbol{23.1}$\quad $\boldsymbol{36.1}$                 & 71.3 \quad  25.0  \quad $\boldsymbol{37.0}$                & 76.3 \quad  8.8\quad   15.8                 \\ \hline
	\end{tabular}}
\end{table*}
space and potential discrimination space and utilize the concated multiple spaces feature to perform ZSL prediction.

Based on the above table, we evaluated ACPSG in detail on three benchmark datasets (AwA2, CUB, aPY). Experimental results show that ACPSG performs well and outperforms some advanced models in some results.

\subsection{Ablation Studies}
To further determine the effectiveness, we conduct the ablation experiments. As shown in Table 2, None refers to the test without using potential discrimination space. The experimental results show that using our generated potential discrimination space can effectively improve the model performance. We have also visualized the predicted attributes, As shown in Fig. \ref{4}, it is easy to find that the distribution distance of each unseen classes in the multiple spaces is larger than the original space, and the class labels are easier to predict, e.g., the distance between Persian cat and rat is much larger than the original space.

\subsection{Generalized Zero-Shot Learning}
In real-world applications, we cannot ensure whether the
test samples are from unseen classes. Thus, the generalized ZSL is more convincing to demonstrate the generalization of models than the conventional ZSL. Hence, we evaluate the performance of ACPSG under the generalized ZSL setting

As shown in Table 3, we tested ACPSG on three benchmark datasets, ACPSG shows a good performance on AwA2 and CUB, But not so good on aPY, we guessed that it was caused by the insufficient fine-grained of the learned distribution. But in general, our model has good generalization capability.

\section{Conclusion}
In this paper, we put forward the concept of attribute correlation in ZSL, and explore the correlation in attribute nodes, it makes attribute nodes are interrelated rather than isolated. To use attribute correlation as a prior knowledge of ZSL, we propose the ACPSG model to make full use of the correlation between nodes. Specifically, our model learns multiple spaces that are more discernible than the original space. Using this method, we integrated attribute correlation into the ZSL model successfully. Besides, we have done a lot of experiments to verify the effectiveness of our model.

In essence, the graph-based approach aims to model the interaction in entities. In our model, classes and attributes are regarded as different nodes in the graph, and edges are used to describe the correlation between the nodes so that the structural information between the various nodes is fully utilized. From all information we discussed, we use the graph convolutional networks to generate a more effective space for potential discrimination.

In reality, we combine the latent discriminating space and the user-defined space into multiple spaces. We train the samples so that the visual features of the samples are mapped into multiple spaces, and the same class is clustered together and distributed reasonably.

There are still many challenges in zero-sample learning. In the future, we will continue to develop ZSL models that based on graphs and attributes to give model better performance and generalization.

\end{document}